\documentclass[twoside,11pt]{article}

\usepackage{blindtext}
\usepackage[preprint]{jmlr2e}

\usepackage{amsmath}
\usepackage{bm}
\usepackage{subfig}
\usepackage{listings}
\usepackage{xspace}

\newcommand{\ie}{\textit{i.e.}\xspace}

\newcommand{\expnumber}[2]{{#1}\mathrm{e}{#2}}
\newcommand{\oned}{\textit{1-d}\xspace}

\newcommand{\threed}{\textit{3-d}\xspace}

\usepackage{lastpage}
\jmlrheading{42}{2266}{1-\pageref{LastPage}}{05/25; Revised M/YY}{M/YY}{21-0000}{E.D.V. Bigarella}

\ShortHeadings{Prevention of Overfitting with a Modified Laplace Operator}{E.D.V. Bigarella}
\firstpageno{1}

\begin{document}

\title{Prevention of Overfitting on Mesh-Structured Data Regressions with a Modified Laplace Operator}

\author{\name Enda D. V. Bigarella \email enda.bigarella@gmail.com \\
       \addr Independent Researcher\\
       Starnberger Weg 7, Gilching, Germany
}

\editor{}

\maketitle

\begin{abstract}%
This document reports on a method for detecting and preventing overfitting on data regressions, herein applied to mesh-like data structures.
The mesh structure allows for the straightforward computation of the Laplace-operator second-order derivatives in a finite-difference fashion for noiseless data.
Derivatives of the \textit{training} data are computed on the original training mesh to serve as a true label of the entropy of the training data.
Derivatives of the \textit{trained} data are computed on a staggered mesh to identify oscillations in the interior of the original training mesh cells.
The loss of the Laplace-operator derivatives is used for hyperparameter optimisation, achieving a reduction of unwanted oscillation through the minimisation of the entropy of the trained model.
In this setup, testing does not require the splitting of points from the training data, and training is thus directly performed on all available training points.
The Laplace operator applied to the trained data on a staggered mesh serves as a surrogate testing metric based on diffusion properties.
\end{abstract}

\begin{keywords}
Diffusion Operator, Overfitting Prevention, Mesh Structure, Surrogate Testing Metric, Gaussian Process Regression
\end{keywords}

\section{Introduction}

The scientific and engineering communities are interested in models that provide good generalisation, and the optimisation of the hyperparameters via the marginal likelihood is often employed as one means for such.
However this is only a common practice and not necessarily a direct outcome of the marginal likelihood definitions, reviewed in what follows, and thus absence of overfitting is not guaranteed.
The current work proposes a diffusion-loss method as an alternative for overfitting detection and prevention for model training.

\subsection{Diffusion Operator and Entropy}
The diffusion operator has widespread use in science as a generalised indicator of entropy metrics.
In the author's background in computational fluid dynamics (CFD) (see \citet{meu_FDM, meu_FVM, meu_Galerkin} for implementations in different discretization schemes), diffusion operators are explicitly or implicitly the basis for detecting numerical oscillations stemming from approximation errors due to discontinuities, such as shock waves (or even strong continuous gradients), or due to inherent numerical characteristics of discretization schemes.
In a different context, that of machine learning, the entropy detection nature of the diffusion term has provided a step improvement in the generation of images \citep{ho2020denoising} in a process termed ``denoising'', which is a particularly interesting term for the current work.

These concepts motivated the proposal of a new metric for model training based on the entropy of the training data. 
The entropy of the training data is herein computed via a modified Laplace operator and applied as a true label.
This term is interpreted as a metric of the ``entropy-in-features'' of the training data.
Testing performed on a staggered mesh bounds the entropy loss with respect to the true label, essentially meaning that a trained model should limit the amount of extra features it determines, and thus less ``noise'' stemming from overfitting.
This concept is further developed into a practical method in Sect.\ \ref{section_method}.

The advantage of this method is that training can be executed on all available training points, and testing is performed on a hierarchical field derived from the training data.
This field developed from a diffusion operator can have different interpretation facets, such as smoothness of curvature, training-introduced noise, or simply overfitting.
Inasmuch the current work focuses on the application to Gaussian Process Regression, the diffusion-loss method is in principle agnostic of the machine learning technique.

\subsection{Gaussian Processes and Marginal Likelihood}
Gaussian Process Regression (GPR) is a powerful asset in engineering applications due to its ability to model complex nonlinear relationships and its inherent confidence interval determination capability.
It has been adopted in surrogate modelling for design optimisation and inverse  problems \citep{forrester2008engineering}, as well as for product data qualification and certification \citep{brunton2020datadrivenaerospaceengineeringreframing}.

For engineering applications, it is typical that the data to be treated is multidimensional, \ie, a representative output parameter of the system depends on many system operational variables.
GPR is particularly suited for this context and as reported by \citet{manzhos2023} might be a superior technique when space dimensionality increases (despite \citet{bengio2005curse}).
In the example to be addressed in this document, a metric of an electric motor efficiency is dependent on three variables, representative of the motor speed, power, and electric feed.
For such types of problems, the system model might be structured following a mesh-like combination of the system variables which are input on a computational model or test bench for the data generation.

It is usually the case that the costs associated with the data generation incur in sparse data especially in cases of high dimensionality.
This is a scenario in which each data point is of extreme value.
It is likely that the fitting must be true to the training data, in the sense that the trained model must exactly represent it.
As a result, a noiseless model training is required, as further discussed in the current work.

A downside of the GPR expressive power, combined with inherent scarcity of training data in multidimensional spaces (\citet{Donoho2000b}), is the difficulty in optimising the model (hyper)parameters for the best generalisation \citep{manzhos2023}.
Direct results of this scenario is a flat objective function landscape, resulting in dependency on initial conditions, and proneness to overfitting.
Noiseless training can reinforce this scenario.

As noted by \citet{Duvenaud2014}, typical GPR kernels are suitable for relatively smooth functions, and discontinuities (including in the first few derivatives) will likely result in ``wiggling''.
However, it might be the case in engineering that the system behaviour is not discontinuous, but it presents changes fast enough to trigger wiggling of a model.
It is an on-going effort in the scientific community to detect and prevent overfitting in data regression.
Many different approaches for GPR (such as \citet{wiggle__cawley2007, wiggle__cawley2010, wiggle__fortuin2020, wiggle__Li_2024, lotfi2023bayesianmodelselectionmarginal}) are reported.
Such techniques generally involve splitting the data into training and testing subsets.
However, for fast changing phenomena such as in Fig.\ \ref{envelope}, the removal of training points is detrimental to the representation of the data.
Furthermore, extracting testing data from the training dataset in already data-limited multidimensional problems might be further detrimental to the quality of the model generalisation.

This complex scenario might be exacerbated by the conventional approach of training the model on the minimisation of the log marginal likelihood (LML).
\citet{lotfi2023bayesianmodelselectionmarginal} provides an interesting discussion on the dichotomy of the expectations between (1) a trained a model based on the marginal likelihood and (2) the quality of the generalisation of the trained model.
The authors point out that ``the marginal likelihood answers the question `what is the probability that a prior model generated the training data?'.''
This is different than answering the question of what model provides the best generalisation, which in fact is the objective for practical application in engineering problems of GP for regression.
This motivated the development of an alternative method to enhance model generalisation as herein proposed.

\section{Gaussian Process Regression}

Regression via Gaussian Process (GP) is a technique \citep{rasmussen2006gaussian} to generate a continuous function $\bm{f}$ and its confidence interval from a sampled dataset.
In the current work, the dataset $\mathcal{M}$ is a $d$-dimensional mesh composed of $n = \prod_{i=1}^d n_i$ samples, with $n_i$ representing the number of points in each of the $d$ dimensions.
The ensemble of the mesh points is given by the system independent variables $X = \{\bm{x}_{(1, 1, ..., 1)},...,\bm{x}_{(n_1, n_2, ..., n_d)}\}$ and the associated observations $\bm{y} = \{y_{(1, 1, ..., 1)},...,y_{(n_1, n_2, ..., n_d)}\}$.

\subsection{Bayesian Conditioning}
A function $\bm{f}$ generated by a GP $\bm{f}(\bm{x}) \sim \mathcal{N} \left( m(\bm{x}), K(X,X) \right)$ has $m(\bm{x})$ as the mean function and $K_{i,j}(X,X) = \mathcal{K}( \bm{x}_{i}, \bm{x}_{j} )$ as the covariance matrix, in this case determined by a covariance kernel $\mathcal{K}(.,.)$ function.
Following the Bayesian methodology, a prior described by the kernel function $\mathcal{K}$ is combined with new data to obtain a posterior distribution.

Given the noise-free training dataset $\mathcal{M}$, the GP prior can be converted into a GP posterior $p(\bm{f}_{*}|X_{*},X,\bm{y})$ to make predictions $\bm{f}_{*}$ at new inputs $X_{*}$.
For test points $X_{*},$ the joint distribution over the function values $\bm{y}$ and $\bm{f}_{*}$ can be written
as
\begin{equation}
\left[ \begin{matrix} \bm{y} \\ \bm{f}_{*} \end{matrix}\right] \sim \mathcal{N} \left(0, \left[ \begin{matrix} K(X,X)&K(X,X_{*}) \\ K(X_{*},X) & K(X_{*},X_{*}) \end{matrix} \right] \right).
\end{equation}
The conditioning of $\bm{f}_{*}$ on $X_{*}$, $X$, and $\bm{y}$ provides the following predictive distribution
\begin{equation}
\begin{aligned}
p(\bm{f}_{*}|X_{*},X,\bm{y}) \sim \mathcal{N}\Big( 
&\ K(X_{*},X) K(X,X)^{-1} \bm{y}, \\
&\ K(X_{*},X_{*}) -  K(X_{*},X) K(X,X)^{-1} K(X,X_{*}) 
\Big).
\end{aligned}
\end{equation}

The leading limitation for GPR application is its computational resource and storage costs.
The training and the prediction with GPR requires the inversion of the kernel matrix, which is of size $n \times n$.
Fortunately the kernel matrix structure allows for a Cholesky decomposition \citep{rasmussen2006gaussian}, which requires $\mathcal{O}(n^{3})$ operations and $\mathcal{O}(n^{2})$ storage costs.

\subsection{Kernel Selection} \label{kernelsel}
The choice of the kernel function $\mathcal{K}$ encodes prior expectation on the function behaviour drawn from the GP.
In general these expectations involve the expressiveness (``flexibility'') and the smoothness of the function, but more advance behaviours such as periodicity, symmetry, or non-stationary properties can also be encoded \citep{rasmussen2006gaussian}.

Probably the most used kernel function for GP is the Squared Exponential (SE) kernel, also referred as the Radial Basis Function kernel, given by:
\begin{equation}
\mathcal{K}_{SE} (x,x^{\prime}) = \sigma^2 ~ \mathrm{exp} \left( -\frac{|x-x^{\prime}|^2}{2 \lambda^2} \right).
\end{equation}
The SE kernel is an attractive choice since it provides infinitely differentiable priors and it has only two hyperparameters, namely:
\begin{itemize}
\item The lengthscale $\lambda$ determines the strength of the correlation of two distant points, which in practice displays itself as the ``wavelength'' of the prior function. The shorter the lengthscale, the more ``wavy'' is the prior space.
\item The variance $\sigma$ determines the strength of features on a prior, in practice the ``amplitude'' of the ``wave'' of a prior function compared to its mean.
\end{itemize}

Another successful kernel option for GP is the Rational Quadratic (RQ) kernel function, given by:
\begin{equation}
\mathcal{K}_{RQ} (x,x^{\prime}) = \sigma^2 \left(1 + \frac{|x-x^{\prime}|^2}{2 \alpha \lambda^2} \right)^{-\alpha}.
\end{equation}
The RQ kernel is widely understood as being equivalent to ensembling many SE kernels with different lengthscales, controlled by the hyperparameter $\alpha$. 
This hyperparameter determines the relative weighing of large-scale and small-scale ratios, with lower $\alpha$ providing a wider spread (more small-scale content) and, in the limit of $\alpha \to \infty$, the RQ is identical to the SE kernel (one single lengthscale).

As noted by \citet{Duvenaud2014}, SE and RQ kernels usually work well for regression of smooth functions.
For functions with discontinuity in itself or in its first derivatives, the trained lengthscale likely results being extremely short and the posterior mean having ``ringing'' effects (that is, drops to zero in between training points).
Even in case of no discontinuities, fast changes still somewhat ``forces'' the lengthscale to being determined by the smallest or fast-changing ``feature'' of the training field, resulting in under- or overfitting issues.

\section{Hyperparameter Optimisation}
As seen in the previous section, the kernel functions introduce trainable hyperparameters.
The generalisation performance of a GP for regression is highly dependent on the choice of such kernel and its hyperparameters.
\citet{bengio2005curse} note that, for multidimensional problems, the need for finding the best balance of kernel (hyper)parameters is further complicated by the so-called bias-variance dilemma.

The minimisation of the LML is widely used for model training in practical applications, and initially reviewed here.
A novel technique based on the minimisation of a modified Laplace diffusion operator is further proposed.

\subsection{Marginal Likelihood Method} \label{LMLsection}
The generalised hyperparameters $\bm\theta$ defining a GP kernel $\mathcal{K}_{\bm\theta}$ can be optimised by typical gradient methods using the LML \citep{rasmussen2006gaussian}, given by
\begin{equation} \label{lmleq}
log~p(\bm{y}|X,\bm\theta) = -\frac{1}{2} \bm{y}^{\top} K_{\bm\theta}^{-1} \bm{y} - \frac{1}{2} log|K_{\bm\theta}| - \frac{n}{2}~log~2\pi.
\end{equation}
The LML can be interpreted as a globally-penalised fit score.
The leading term measures the data fit score and the mid term is a complexity penalisation score.
A model trained by the minimisation of the LML should provide a good data fit with the lowest complexity, in principle.
The minimisation of the LML requires resource-demanding matrix operations which can be a concern especially for larger datasets.

\citet{lotfi2023bayesianmodelselectionmarginal} thoroughly review the use of the LML for model training.
As summary of positive points, and possibly the reason for its wide adoption as hyperparameter optimisation: it automatically favours the most constrained model (``Occam's Razor''), likely outperforming other approaches such as cross-validation; furthermore, the gradients of the LML with respect to the hyperparameters can be analytically computed at the training data points, rendering it computationally cost effective where cross-validation would suffer from a curse of hyperparameter dimensionality.
In practice the LML minimisation can be an effective approach for hyperparameter optimisation.

However, ``marginal likelihood is not generalization.'' 
In model selection the intent is to evaluate posteriors and not (prior) hypothesis testing, and LML is by its nature however mostly tightly linked to the latter.
The LML optimisation can be prone to both underfitting and overfitting, and not necessarily provide the best generalisation after training.

In the current work, the minimisation of the LML is performed with the Python-library GPFlow $minimize$ routine, employing the limited-memory BFGS (L-BFGS-B, \citet{LBFGSB}) algorithm.
A great feature of the GPFlow minimiser is its computation of the gradients of the LML with respect to the hyperparameters by employing a back-propagation scheme \citep{Goodfellow-et-al-2016}, rendering the method more robust for practical applications.

\subsection{Diffusion-Loss Method} \label{section_method}
In the CFD community, the identification and prevention of numerical oscillations is a fundamental concern and a subject of basic teaching \citep{jameson_2012}.
These oscillations mainly stem from two mechanisms: discontinuities or fast change of flow properties, such as shock waves, leading to the ``Gibbs'' phenomenon \citep{fourieranalyses}; or odd-even decoupling due to first-order numerical differentiation based on centred stencils \citep{jameson_2012,TANG2018404}.

\subsubsection{Oscillation Sensor}
\begin{figure}
	\centering
	\includegraphics[width=10cm]{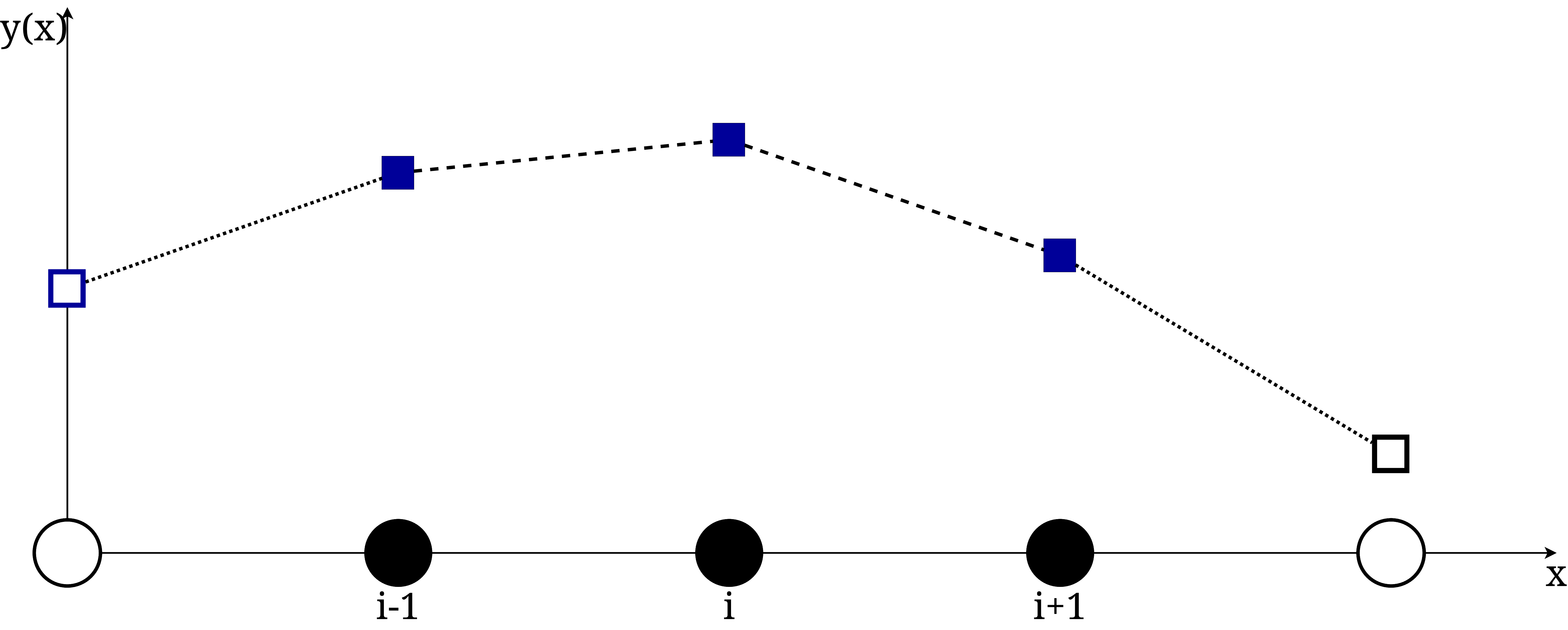}
	\caption{A one-dimensional mesh: the function values are shown by squares and the mesh nodes by circles. The filled symbols show the nodes used to compute the diffusion sensor for the $i$-th element.}
	\label{normal1d}
\end{figure}

One method in CFD for identifying oscillations is to use a measure of the diffusion of the pressure field \citep{tv94, scalabrin_unstructured_2005}, represented by a modified Laplace operator $\nabla$.
This indicator can also subsequently be used to \textit{prevent} oscillations by scaling oscillation control functions.
Expanding on the so-called ``undivided Laplacian'' from the CFD literature, this sensor in a \oned finite-difference context is computed as follows
\begin{equation} \label{truescore1d}
\nabla y_i = \frac{1}{\Delta^2} \frac{\big| y_{i+1} - 2 y_{i} + y_{i-1} \big|}{y_{i+1} + 2 y_{i} + y_{i-1}},
\end{equation}
for the $i$-th point of the mesh in Fig.\ \ref{normal1d}, with $\Delta$ representing the mesh spacing, and $y$ a positive property.
In the currently proposed method, this diffusion sensor applied to the \textit{training} data serves as the \textit{true label}.

\begin{figure}
	\centering
	\includegraphics[width=10cm]{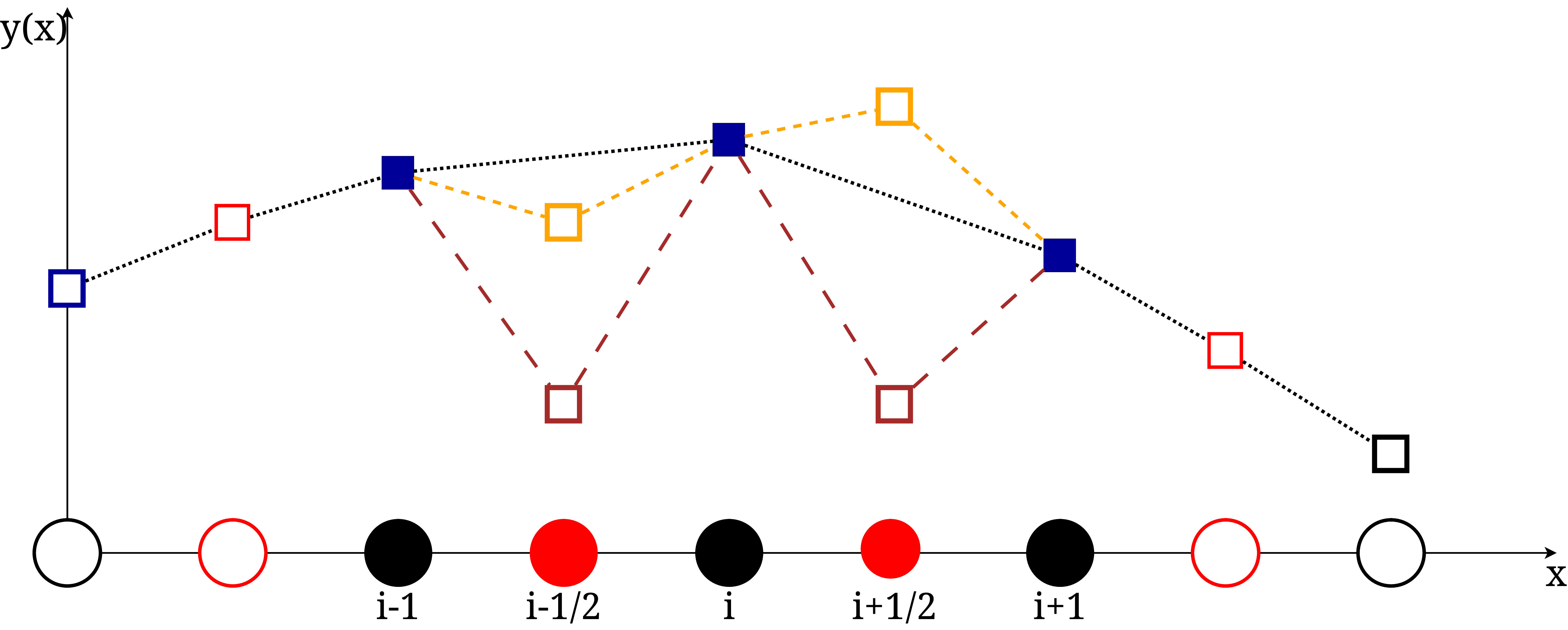}
	\caption{The original training mesh shown as black nodes and the created staggered nodes shown in red, with half indices. Brown and orange squares show possible oscillations around node $i$.}
	\label{staggered1d}
\end{figure}

For a \textit{trained} noiseless model, its prediction at the training points should exactly recover the training data.
The diffusion sensor applied to these points should exactly recover the training true label and would be insensitive to actual overfitting.
Therefore, in order to create a suitable \textit{testing} drive, staggered nodes centred between the training nodes are created, shown in red in Fig.\ \ref{staggered1d}.
An updated diffusion sensor $\tilde\nabla$ considering these staggered nodes, indexed by half indices, is here proposed
\begin{equation} \label{staggeredlaplacian}
\tilde\nabla y_i = \frac{1}{3~(\Delta/2)^2} \frac{\big| \hat{y}_{i+1} - \hat{y}_{i+\frac{1}{2}} - \hat{y}_{i-\frac{1}{2}} + \hat{y}_{i-1} \big|}{\hat{y}_{i+1} + \hat{y}_{i+\frac{1}{2}} + \hat{y}_{i-\frac{1}{2}} + \hat{y}_{i-1}},
\end{equation}
with $\hat{y}$ representing \textit{predicted} function values, $\hat{y}_i = f(x)_i$, with $\hat{y} = y$ expected at the training points.
This updated diffusion sensor is roughly the average of the three possible 3-point stencils centred at ${i-\frac{1}{2}}$, $i$, and ${i+\frac{1}{2}}$, explaining the reason for the 3 in the denominator.
Since we are taking these Laplace operators on the half-points of the staggered points, the mesh spacing is half of the original mesh.

This stencil is able to identify the ``ringing'' (in brown) and ``wiggling'' (in orange) features in Fig.\ \ref{staggered1d} around the central node $i$.
In considering this ability when applied to a staggered mesh, deviations between a testing sensor (Eq.\ \ref{staggeredlaplacian}) and the true score (Eq.\ \ref{truescore1d}) can thus be used as loss in model training.

This methodology allows for the creation of a hierarchical testing metric from the training data field.
Hence, splitting a limited training dataset into testing sets is not required.
Training is performed on the complete available dataset and offers best chance for generalisation. 
The proposed method requires a staggered mesh to be built, function predictions on its nodes to be executed, and the sensor in Eq.\ \ref{staggeredlaplacian} to be computed, which is straightforward and computationally amenable.

\subsubsection{Multidimensional Implementation} \label{mdlaplaciansection}
\begin{figure}
	\centering
	\includegraphics[width=5cm]{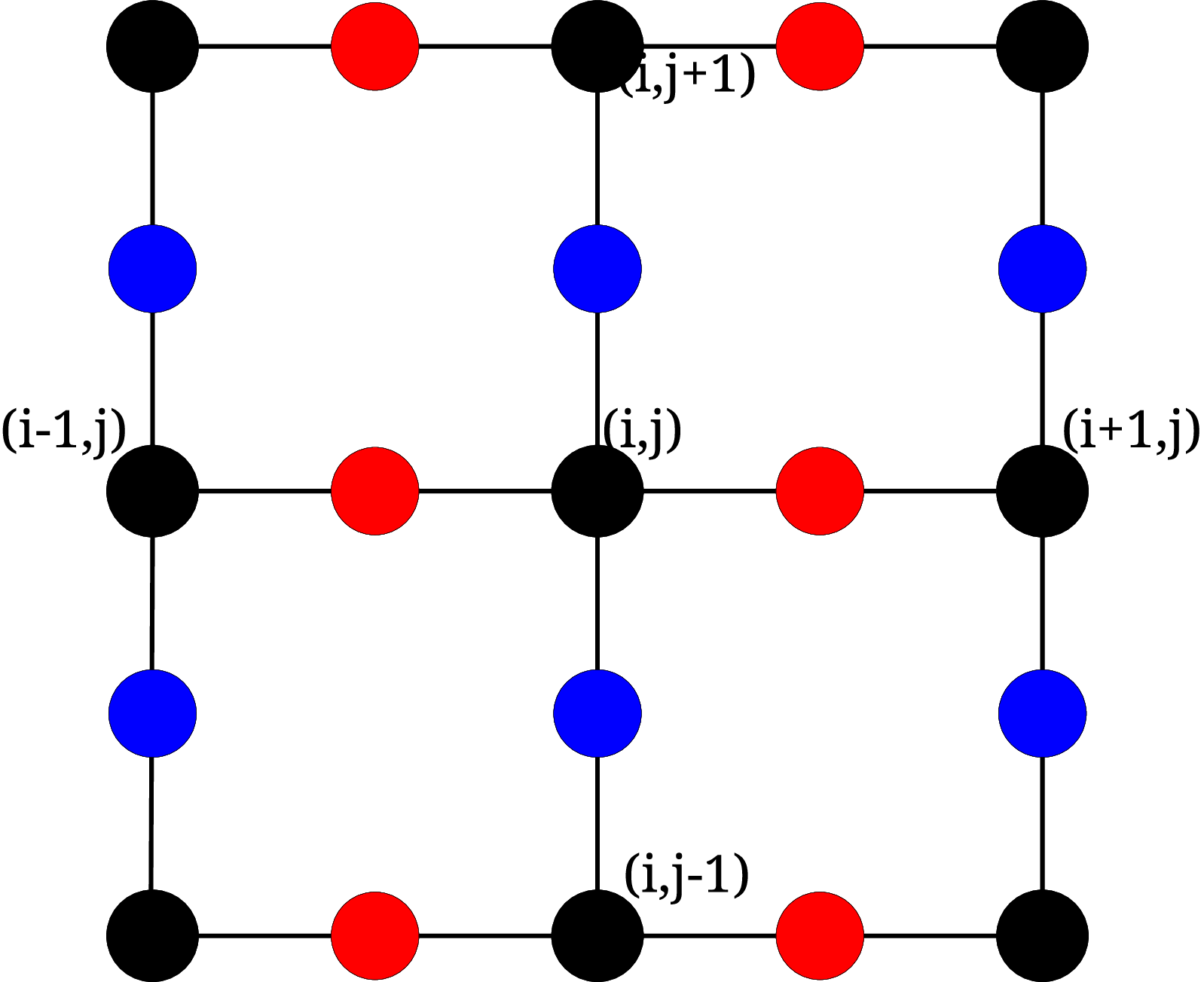}
	\caption{The direct application of the one-dimensional staggering strategy to a bidimensional problem requires two independent staggered meshes based on the edges for each direction of the problem, in blue and red.}
	\label{staggered2ddirect}
\end{figure}

The creation of a staggered mesh in one dimension as shown in Fig.\ \ref{staggered1d} is straightforward since a central node only has a single neighbour to either its side.
On a multidimensional case, for instance a bidimensional problem, the central node has neighbours to the west and east, along $\bm{x}_{(1,0)}$, and north and south, along $\bm{x}_{(0,1)}$.
The creation of staggered points would require independent points along each direction, as shown in Fig.\ \ref{staggered2ddirect}.
This is not a problem \textit{per se} but requires the management of two independent meshes (red and blue nodes in Fig.\ \ref{staggered2ddirect}), doubling the amount of predictions per training cycle.

\begin{figure}
  \centering
  \subfloat[Bidimensional mesh.]{
      \includegraphics[width=0.4\textwidth]{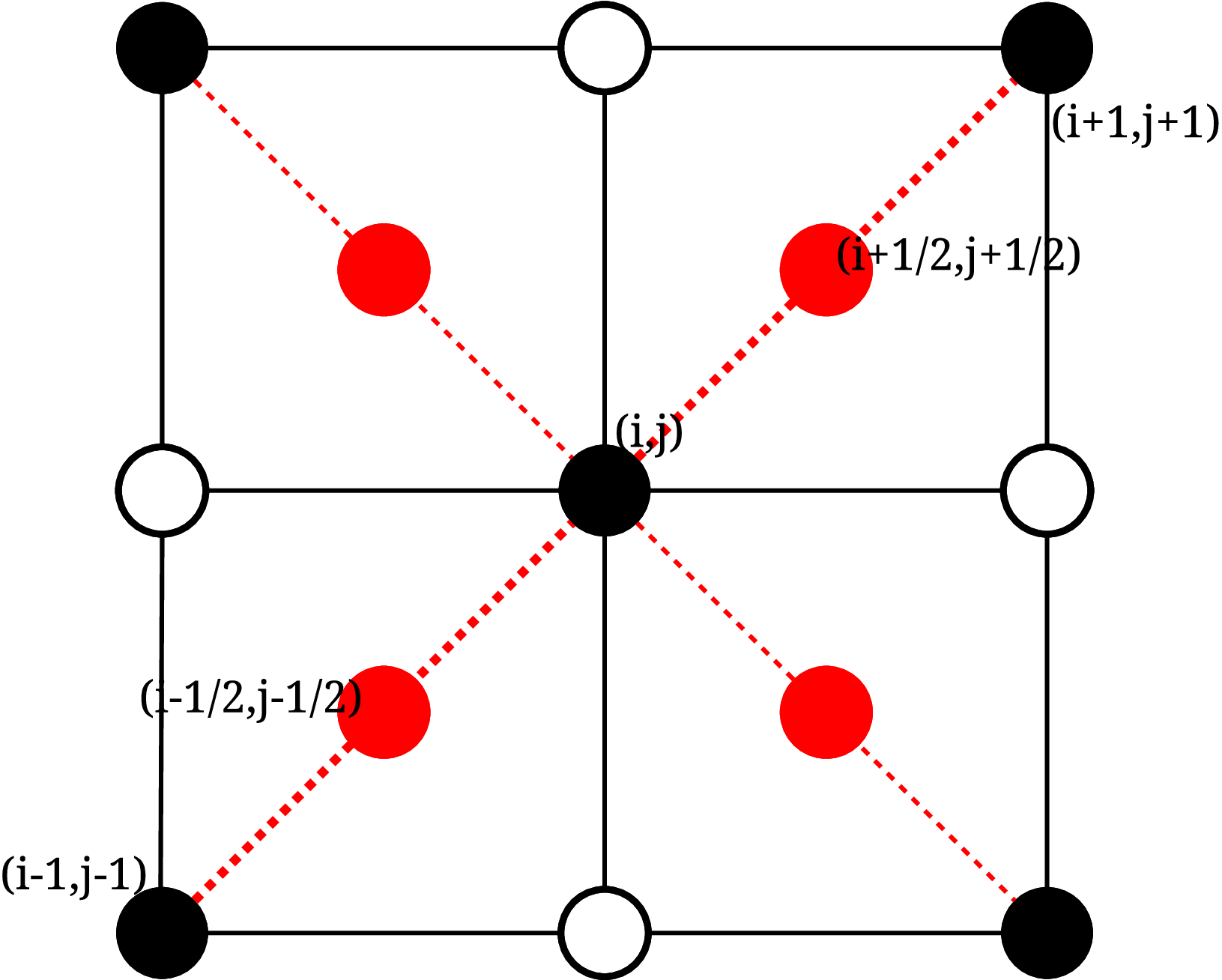}}
  \hfill
  \subfloat[Three-dimensional mesh.]{
      \includegraphics[width=0.55\textwidth]{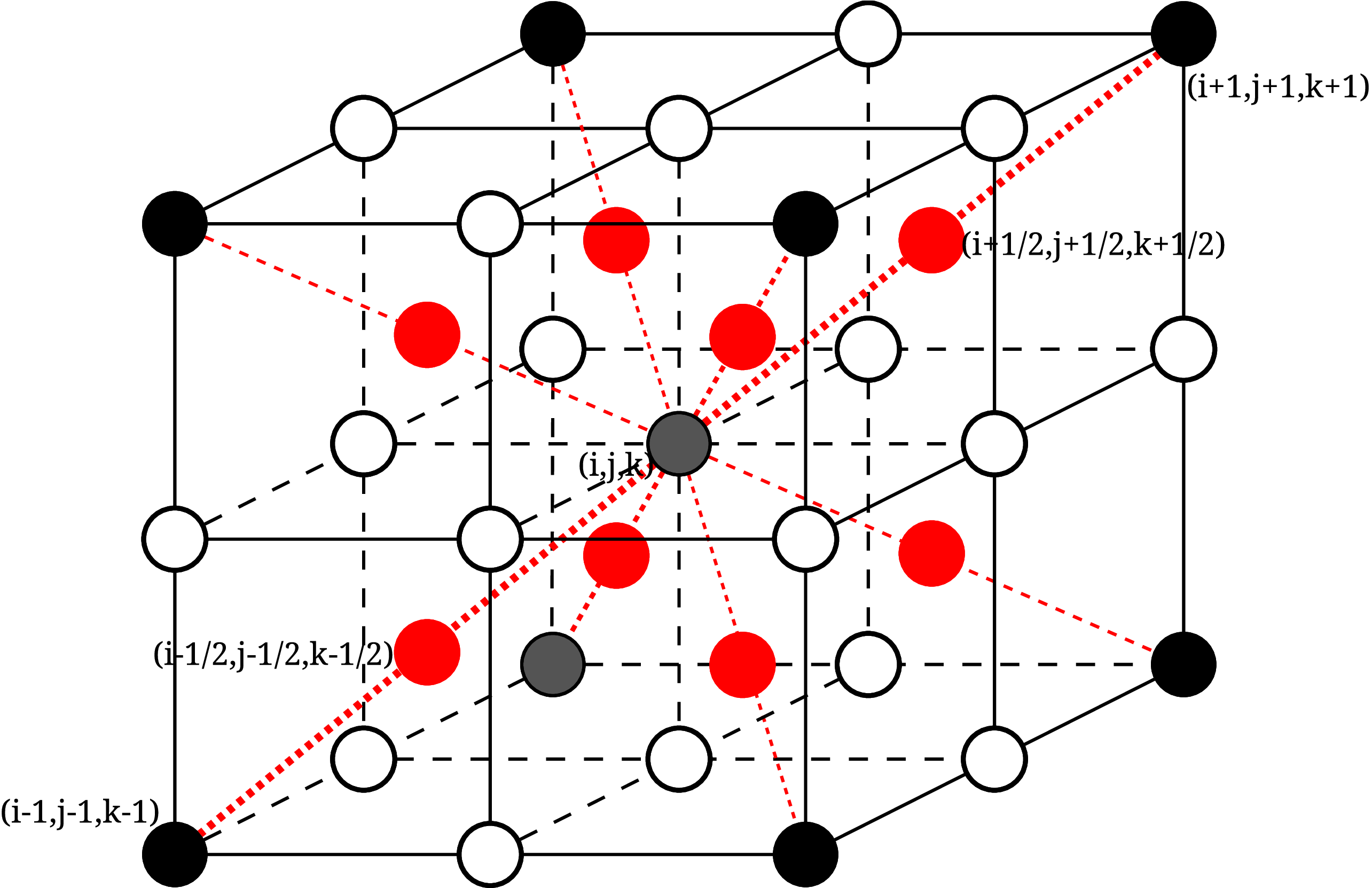}}
  \caption{The original multidimensional training mesh shown as black points and the created cell-centred staggered points shown in red, with half indices. The diffusion sensors are computed along the centre-crossing diagonals shown in red. Filled symbols are used for the sensor computations for the central node ($(i,j,k)$ in \threed, for instance).}
  \label{staggereddiag}
\end{figure}

In considering that the intent of this work is not to create the exact Laplace operator, rather a practical metric out of its concept, an approach based on a single staggered mesh built on centroids of the original mesh cells is proposed.
Resulting staggered mesh examples for a bidimensional and a three-dimensional case are shown in Fig.\ \ref{staggereddiag}.
The mesh dataset $\mathcal{M}$ thus generates a cell-centred staggered mesh $\mathcal{S}$ of $d$-dimension, composed of $s = \prod_{i=1}^d (n_i-1)$ points.
The ensemble of the staggered mesh points is given by the coordinates $X_s = \{\bm{x}_{(1/2, 1/2, ..., 1/2)},...,\bm{x}_{(n_1-1/2, n_2-1/2, ..., n_d-1/2)}\}$.

In this approach, we are required to use the centre-crossing diagonals (composed of pairs of antipodal points) as an \textit{ad-hoc} approach to compute Laplace operators.
The \textit{staggered} Laplace-operator diffusion sensor written for the centre-crossing diagonal $p$ of the node $(i,j,k)$ of the \threed mesh in Fig.\ \ref{staggereddiag} is given by
\begin{equation} \label{staggeredmdlaplacian}
\tilde\nabla y_{i,j,k}^{p} = \frac{1}{3~(\Delta/2)^2} \frac{\big| y_{i+1,j+1,k+1} - y_{i+\frac{1}{2},j+\frac{1}{2},k+\frac{1}{2}} - y_{i-\frac{1}{2},j-\frac{1}{2},k-\frac{1}{2}} + y_{i-1,j-1,k-1} \big|}{y_{i+1,j+1,k+1} + y_{i+\frac{1}{2},j+\frac{1}{2},k+\frac{1}{2}} + y_{i-\frac{1}{2},j-\frac{1}{2},k-\frac{1}{2}} + y_{i-1,j-1,k-1}},
\end{equation}
and for the true label
\begin{equation} \label{mdlaplacian}
\nabla y_{i,j,k}^{p} = \frac{1}{\Delta^2} \frac{\big| y_{i+1,j+1,k+1} - 2 y_{i,j,k} + y_{i-1,j-1,k-1} \big|}{y_{i+1,j+1,k+1} + 2 y_{i,j,k} + y_{i-1,j-1,k-1}}.
\end{equation}
The complete \threed diffusion sensors for the node $(i,j,k)$ are given by the summation over all centre-crossing diagonals resulting in
\begin{equation}
\nabla y_{i,j,k} = \sum_{p=1}^{4} \nabla y_{i,j,k}^{p}, \, \tilde\nabla y_{i,j,k} = \sum_{p=1}^{4} \tilde\nabla y_{i,j,k}^{p}.
\end{equation}
Note that the diffusion sensors are only computed at the interior of the original mesh, where the complete stencil is available.
Check Appendix \ref{appA} for an example of the implementation in \threed.

For a uniform mesh in two dimensions, the diagonal Laplace operator exactly recovers the Laplace operator computed along the coordinate directions.
For higher dimensions, the number of diagonals is larger than the number of dimensions and would require an increasingly number of computations for the proposed diffusion sensor.
The number of centre-crossing diagonals for dimension $n$ is $2^{n-1}$.
However, this approach is still more resource friendly than managing $n$ independent staggered meshes for each direction as the result of the immediate application of the \oned definition in Fig.\ \ref{staggered2ddirect}.
It is worth noting that even though we are summing more diagonals than coordinates, this is not a problem for the  computation of the loss since we are fairly comparing similar computations with the same number of diagonals.
Nonetheless, for cases of very high dimensionality, the author suggests selecting the first $d$ diagonals with the highest diffusion sensor scores, and use these for building the diffusion loss, instead of the complete set of centre-crossing diagonals.

\subsubsection{Minimisation of the Losses}
There are essentially two metrics that compose the total loss $\mathcal{L}$ to be minimised in the proposed approach, \ie, the loss of the training data and the loss of the diffusion sensor, resulting in
\begin{equation} \label{daloss}
\mathcal{L} = \beta_1~\mathrm{RMSE}_{training} + \beta_2~\mathrm{RMSE}_{diffusion},
\end{equation}
where $\beta_1$ and $\beta_2$ are weights, here considered 1 for both, and the conventional root mean square errors $\mathrm{RMSE}$ written as
\begin{equation}
\begin{aligned}
\mathrm{RMSE}_{training} = \sqrt{\frac{1}{n} \sum_{e=1}^n (y_e - f(x_e))^2 }, \\
\mathrm{RMSE}_{diffusion} = \sqrt{\frac{1}{n^*} \sum_{e=1}^{n^*} (\nabla y_{e} - \tilde\nabla y_{e})^2 }.
\end{aligned}
\end{equation}
where index $e$ represents an element of the mesh, and $n^* = \prod_{i=1}^d (n_i-2)$ accounts for the interior nodes of the training mesh.

The adoption of the RMSE seems to provide the best balance overall between the two losses.
The losses are minimised employing the COBYQA algorithm \citep{razh_cobyqa} available in the Python-library SciPy $minimize$ routine.
COBYQA is a derivative-free optimiser and a generally superior choice for the objective function landscape resulting from the loss $\mathcal{L}$ characteristics, especially in cases where ``ringing'' is occurring and gradient-based methods get stuck due to a flat gradient field.

\section{Results and Discussions}

\begin{figure}
	\centering
	\includegraphics[clip, trim=3cm 3cm 3cm 3cm, width=11cm]{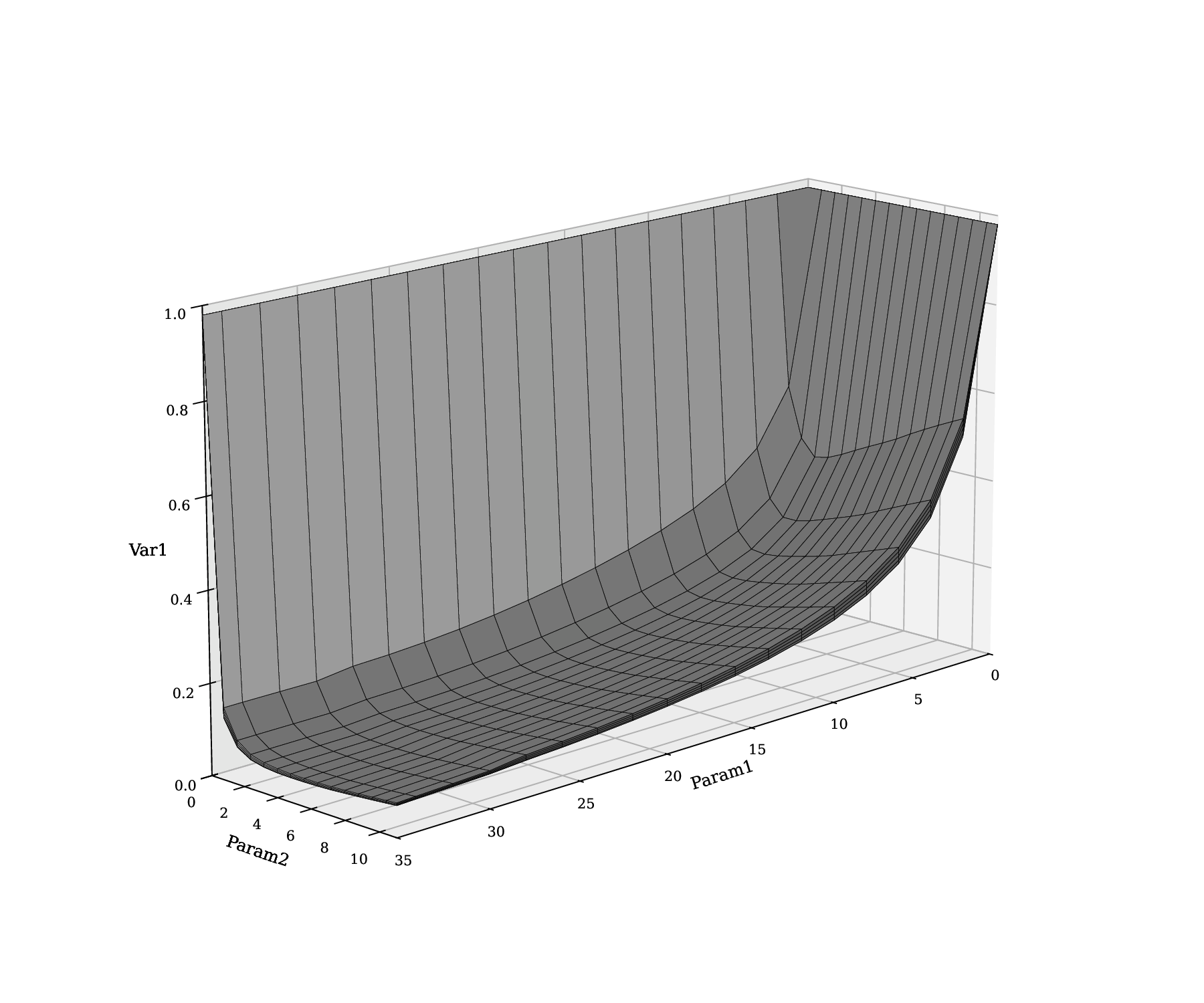}
	\caption{Wireframe view of the mesh showing the two main dimensions, with the third dimension shown as the ``thickness'' of the wireframe.}
	\label{envelope}
\end{figure}

The model problem described in the introduction is composed of a \threed mesh of \textit{(param1, param2, param3)} = $(19 \times 15 \times 5)$ points, shown in Fig.\ \ref{envelope}.
The problem shows a fast change of behaviour along $param1$ and an even faster one with $param2$, which combine at one corner of the operational envelope of the motor.
The behaviour with $param3$ shows as the ``thickness'' of the wireframe, and it changes with both other parameters.
A nonlinear feature can also be observed approaching the maximum limits of the $param1$ range.

The dataset is normalised so as to keep unitary mesh spacing in all dimensions, which brings advantages and avoids skewness in the computation of the diffusion terms.
The system parameter $var1$ is naturally in the range of $[0:1]$ and kept as is.
For general application of the method, it is necessary that the training function values $y$ (Sect.\ \ref{section_method}) are normalized to a positive range.

A numerical study is performed with both SE and RQ kernels, \textit{initialised} with different combinations of hyperparameters, to asses their effect on the \textit{final} ``optimised'' hyperparameters and respective trained data.
For the sake of conciseness and since modifications to the following parameters do not change the conclusions herein presented, a single kernel is used and the variance is fixed to $\sigma = 0.15$, which was preliminarily determined to give best representation at the flatter boundaries of the domain.
The lengthscale $\lambda$ is initialised with values $[0.01, 0.01, 0.5, 1, 5, 10]$ and the RQ $\alpha$, in the range $[0.005, 0.05, 0.5, 5, 10, 50]$.
The numerical experiment is implemented in Python using the GPFlow library \citep{GPflow2017}.

The motivation and rationale for a noiseless training for the current exercise is initially provided.
In the sequence, the optimised results for the LML and the diffusion losses methods are presented and discussed.

\subsection{Noiseless Training}
As previously stated, for the current study the trained data must exactly reproduce the training information, thus termed a noiseless training.
The explanation for this choice stems from the author's background in aircraft certification.
Typically a certified method to expand data for operation must ``go through'' the \textit{certified data}, considered the source of truth since this data was measured in certified tests and signed off.
Naturally these experimental sources are subject to ``scatter'', but generally the approach to develop the certified data from this, is to have a few repetitions of the same test point performed and to \textit{average} them out \citep{faa_ac_29-2c,faa_ac_25-7d}.
This averaged data point, from those repetitions, is thus the certified data.
The process is repeated for other operational conditions and the whole dataset is built from this.
In the case of data from accepted computational methods, noise is not expected.
Therefore, in all cases, the data expansion provided by a fitting process such as a GPR must recover the certified data used for training.

A noiseless training is generally more challenging and is an additional reason for this choice in the current exercise.
For the LML hyperparameter optimisation, a Gaussian \textit{likelihood} is set up with a fixed $\expnumber{1}{-10}$ noise, to guarantee a noise-free fitting.
For the diffusion loss method, such mathematically strong enforcement is not guaranteed but is nonetheless an expected outcome from the inclusion of $\mathrm{RMSE}_{training}$ in Eq.\ \ref{daloss}, and is further assessed in this work.

\subsection{LML Minimisation Results}
\begin{figure}
  \centering
  \subfloat[SE kernel.]{
      \includegraphics[width=0.475\textwidth]{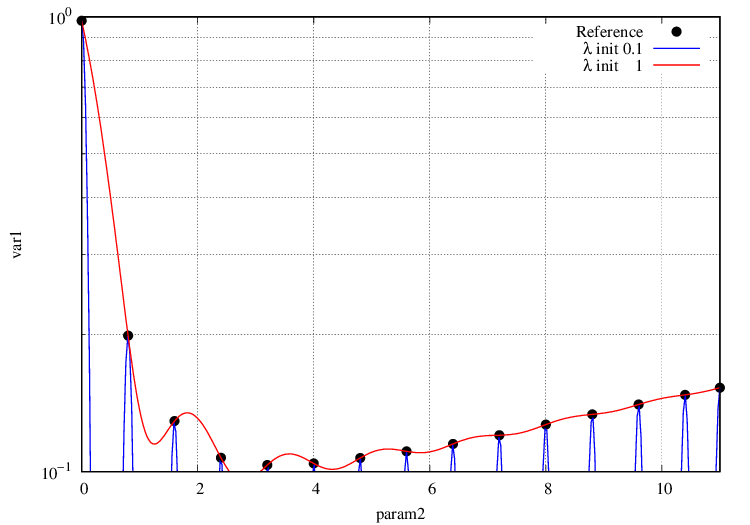}}
  \hfill
  \subfloat[RQ kernel.]{
      \includegraphics[width=0.475\textwidth]{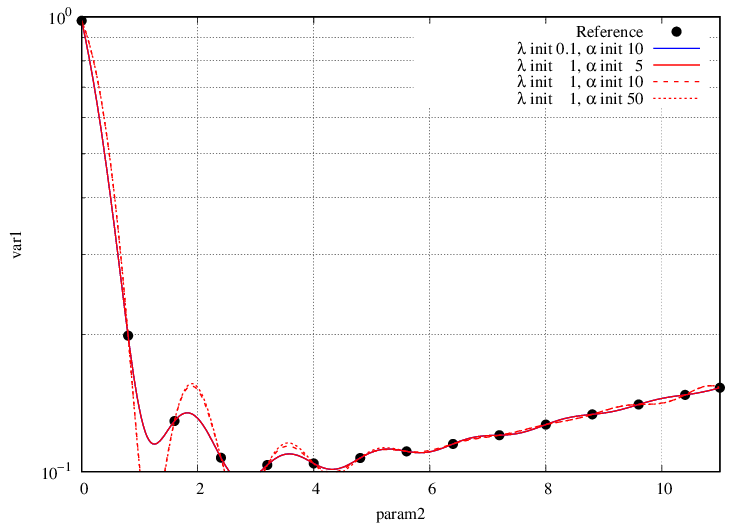}}
  \caption{Trained data based on the LML minimisation sliced at $(param1, param3) = (14, 0.7)$ for different initial values of hyperparameters.}
  \label{result_lml_loss}
\end{figure}

Figure \ref{result_lml_loss} shows the most representative results from the numerical experiment developed according to the minimisation of the LML, described in Sect.\ \ref{LMLsection}.
Overfitting indications such as ``wiggling'' or ``ringing'' are clearly observed in all cases.
With the smaller initial length scales, the LML minimisation can get stuck possibly due the ``ringing'' effect incurring in flatness of the LML gradients.
The RQ kernel does not present the same tendency at lower lengthscales and managed to ``get out of the well'', but shows some dependency on the hyperparameter $\alpha$.

\begin{figure}
  \centering
  \subfloat[SE kernel.]{
      \includegraphics[width=0.475\textwidth]{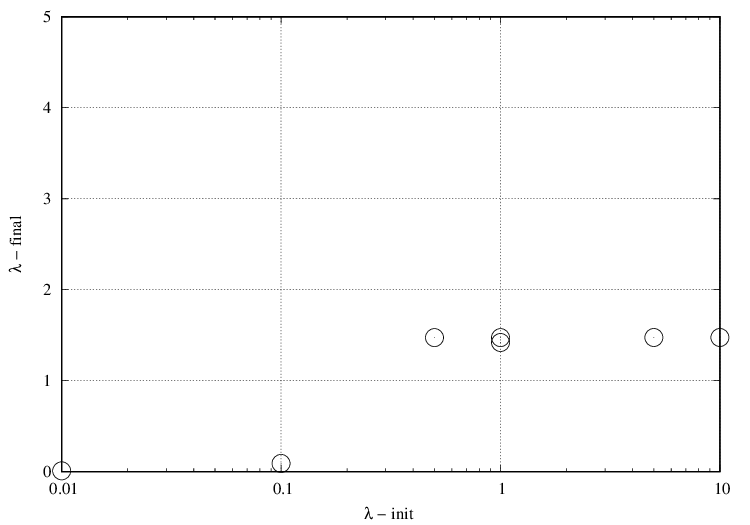}}
  \hfill
  \subfloat[RQ kernel.]{
      \includegraphics[width=0.475\textwidth]{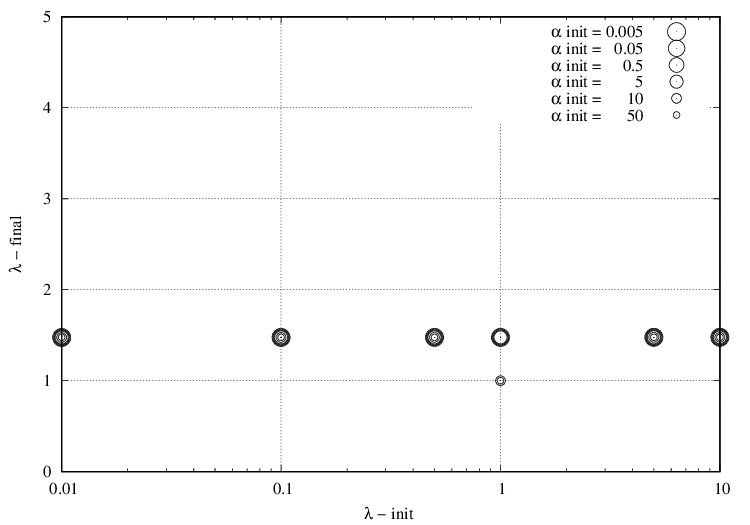}}
  \caption{Final lengthscales for different initial values of hyperparameters from the LML minimisation.}
  \label{result_loss_lengths}
\end{figure}

\begin{figure}
  \centering
  \subfloat[SE kernel.]{
      \includegraphics[width=0.475\textwidth]{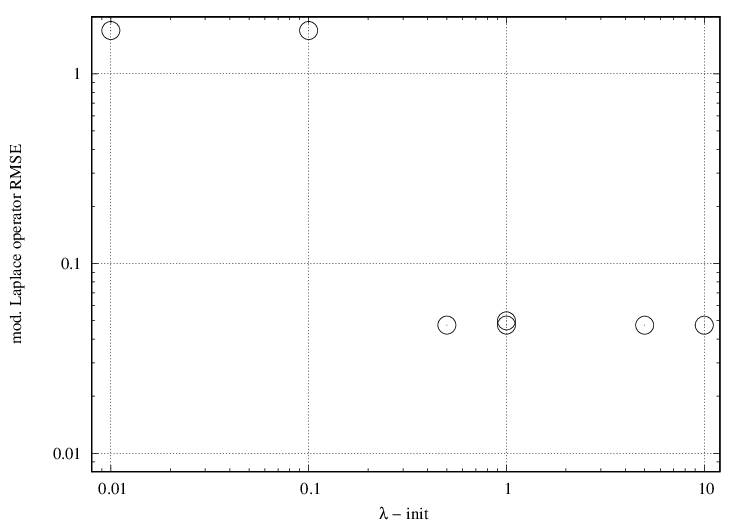}}
  \hfill
  \subfloat[RQ kernel.]{
      \includegraphics[width=0.475\textwidth]{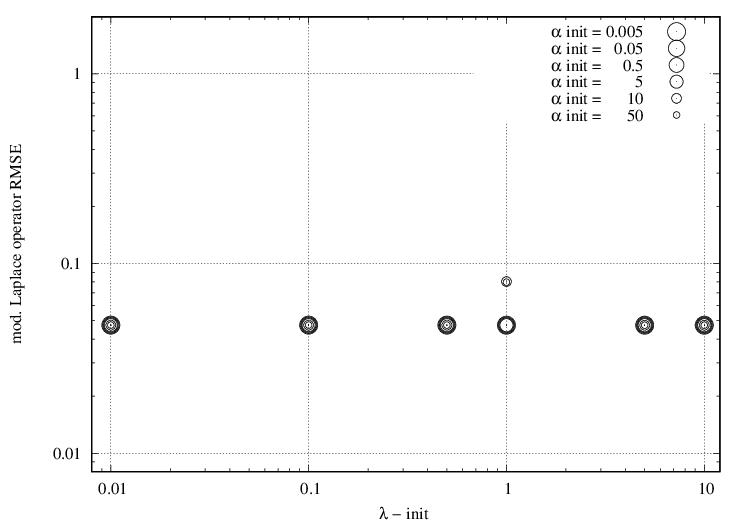}}
  \caption{Final diffusion losses for different initial values of hyperparameters from the LML minimisation.}
  \label{result_loss_Lrmse}
\end{figure}

The final lengthscales obtained from the complete numerical experiment are plotted in Fig.\ \ref{result_loss_lengths}.
The respective diffusion losses (Sect.\ \ref{mdlaplaciansection}) are shown in Fig.\ \ref{result_loss_Lrmse}.
The SE result subject to the ``ringing'' effect in Fig.\ \ref{result_lml_loss} shows a consistently higher diffusion loss.
The RQ kernel presents a more complex optimisation landscape with the optimisation of $\alpha$ but generally converged to the same most-preferred lengthscale of the SE kernel.
In all cases, $\alpha$ converged to values in the range of 500--1000 (not shown), thereby approximating the RQ kernel to the SE (Sect.\ \ref{kernelsel}) and not allowing for reaping any additional feature from it.
The diffusion losses of the most-preferred lengthscales are very similar as expected, as seen in Fig.\ \ref{result_loss_Lrmse}.

\subsection{Diffusion-Loss Minimisation Results}
\begin{figure}
  \centering
  \subfloat[SE kernel.]{
      \includegraphics[width=0.475\textwidth]{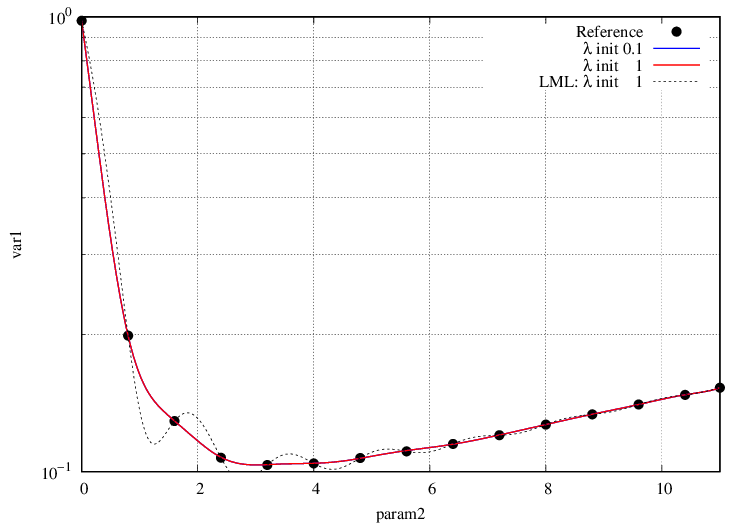}}
  \hfill
  \subfloat[RQ kernel.]{
      \includegraphics[width=0.475\textwidth]{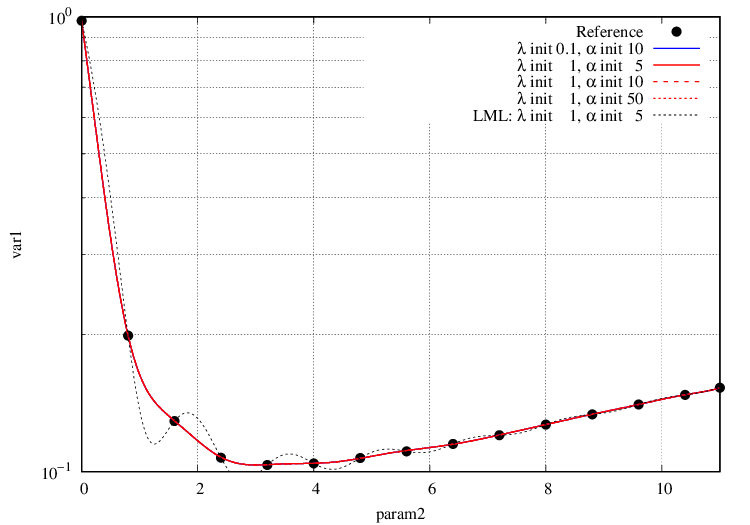}}
  \caption{Trained data based on the diffusion-loss minimisation sliced at $(param1, param3) = (14, 0.7)$ for different initial values of hyperparameters. Equivalent LML results are shown for reference.}
  \label{result_lapl_loss}
\end{figure}

The previously executed numerical experiment is replicated with the diffusion-loss minimisation method proposed in Sect.\ \ref{mdlaplaciansection}, with an initial search radius of 3 for the COBYQA optimiser.
Similarly to Fig.\ \ref{result_lml_loss}, slices of the respective trained data results are shown in Fig.\ \ref{result_lapl_loss}.
These results are indicative of the reduction of overfitting with the proposed diffusion-loss method.

\begin{figure}
  \centering
  \subfloat[SE kernel.]{
      \includegraphics[width=0.475\textwidth]{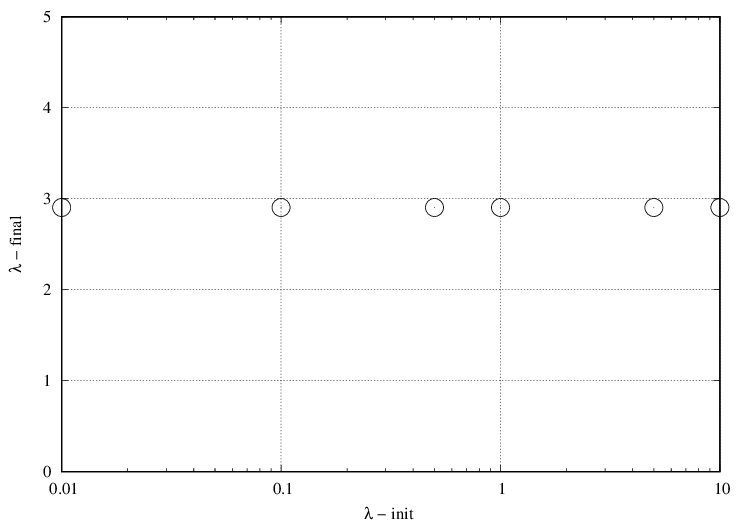}}
  \hfill
  \subfloat[RQ kernel.]{
      \includegraphics[width=0.475\textwidth]{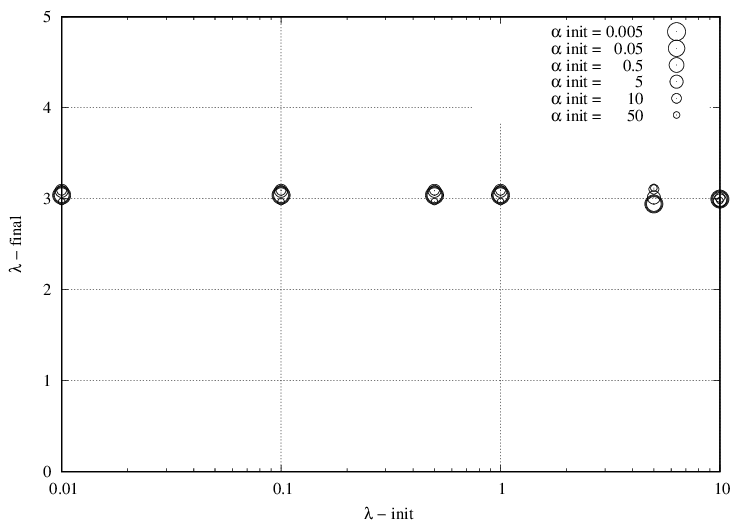}}
  \caption{Final lengthscales for different initial values of hyperparameters from the diffusion-loss minimisation.}
  \label{result_loss_lengths_lapl}
\end{figure}

\begin{figure}
  \centering
  \subfloat[SE kernel.]{
      \includegraphics[width=0.475\textwidth]{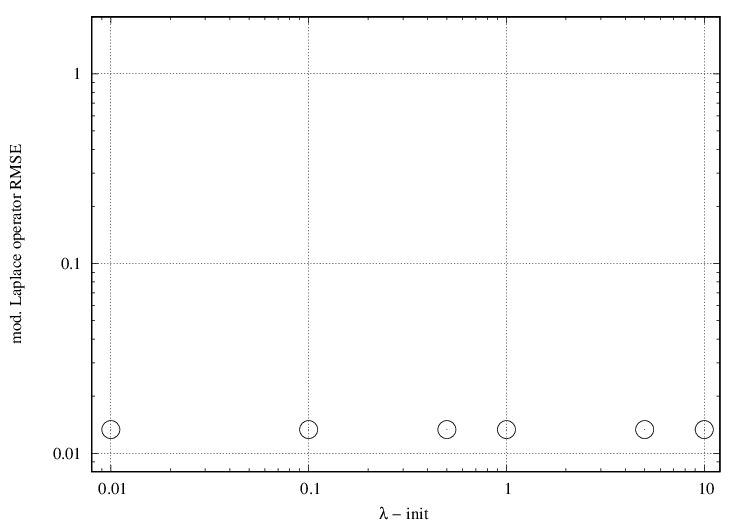}}
  \hfill
  \subfloat[RQ kernel.]{
      \includegraphics[width=0.475\textwidth]{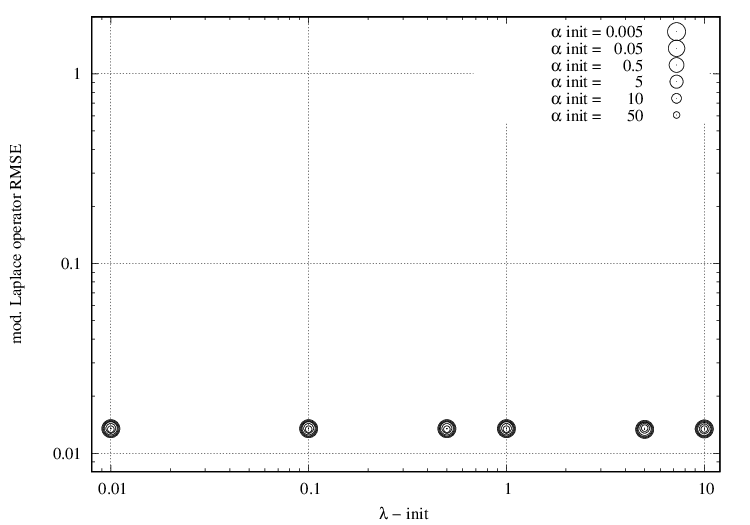}}
  \caption{Final diffusion losses for different initial values of hyperparameters from the diffusion-loss minimisation.}
  \label{result_loss_Lrmse_lapl}
\end{figure}

The final lengthscales obtained from the complete numerical experiment are plotted in Fig.\ \ref{result_loss_lengths_lapl}.
The respective diffusion losses are shown in Fig.\ \ref{result_loss_Lrmse_lapl}.
The proposed diffusion-loss method substantially decreases the dependency with initial parameters.
Consistent 3-fold lower diffusion losses compared to those obtained with the most-preferred LML lengthscales are observed.
The RQ kernel is still prone to local minima, however these are very close to each other in terms of the diffusion-loss metric.
All cases generally converge to a slightly different lengthscale with final $\alpha$ (not shown) in the range of 50--115 (a few at 250), nonetheless the resulting diffusion losses are very similar to each other, as seen in Fig.\ \ref{result_loss_Lrmse_lapl}.

\begin{figure}
  \centering
  \subfloat[SE kernel, initial $\lambda = 0.01$.]{
      \includegraphics[width=0.475\textwidth]{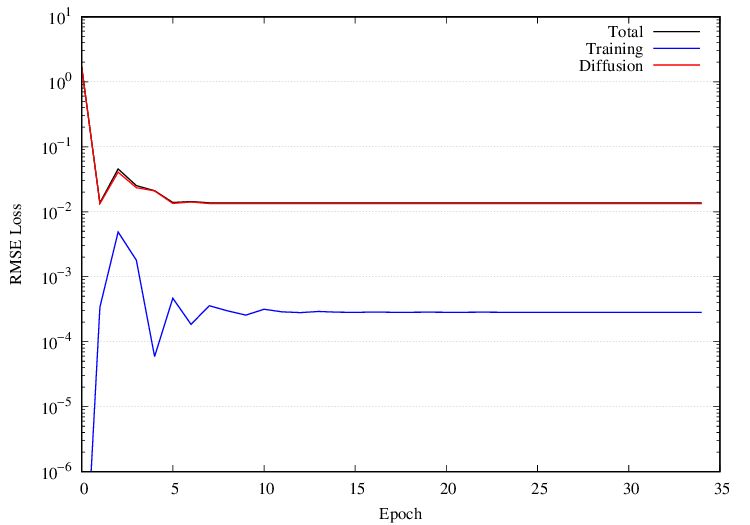}}
  \hfill
  \subfloat[RQ kernel, initial $\lambda = 0.01$, $\alpha = 0.005$.]{
      \includegraphics[width=0.475\textwidth]{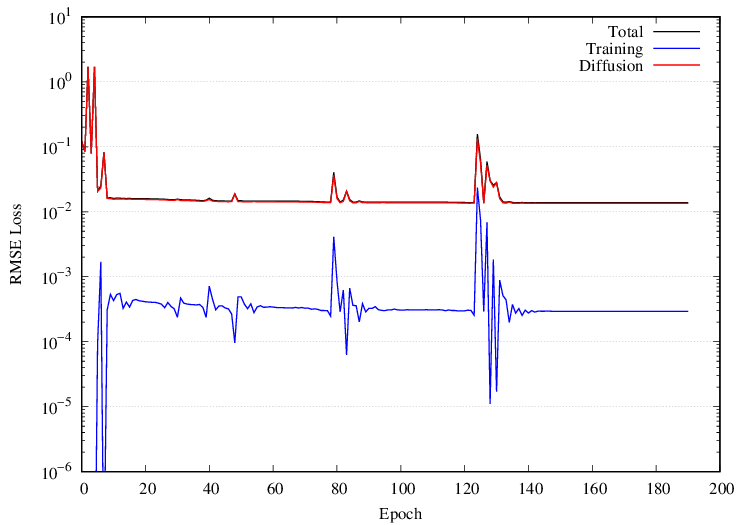}}
  \caption{Two most demanding training- and diffusion-loss convergence histories.}
  \label{convergence_lapl_loss}
\end{figure}

It is interesting to remark that the ability of the method to avoid getting stuck in lower lengthscales can also partially be attributed to the properties of the COBYQA optimisation method.
However, the LML minimisation did not improve with the adoption of this optimisation algorithm.
This observation can be interpreted as a favourable change towards a less flat objective function landscape with the diffusion-loss definitions.
The convergence histories for the most challenging optimisations, initialised with the lowest lengthscale in the study range, are shown in Fig.\ \ref{convergence_lapl_loss}.
As previously stated, the RQ additional hyperparameter adds complexity to the objective function and more epochs are required for the optimisation.

In all cases, the $\mathrm{RMSE}_{training}$ loss reaches a floor of $\expnumber{2.9}{-4}$ as seen in Fig.\ \ref{convergence_lapl_loss}, whereas the equivalent LML results are consistent at $\expnumber{1.3}{-7}$ (not plotted here).
The diffusion-loss minimisation clearly does not reach the same level of accuracy on the training data.
This higher $\mathrm{RMSE}_{training}$ can be assumed as the ``price to pay'' for the reduction of the overfitting, and may be evaluated if adequately low for practical purposes.
The author recalls that the weighing parameters in the loss definition in Eq.\ \ref{daloss} can be used for further adjustments, if necessary.

It might be argued based on this behaviour that the diffusion-loss minimisation acts as a noise variance and similar results could be achieved by simply modelling a noisy regression.
This assertion is not principled since the noise acts as a stochastic process whereas the minimisation of the diffusion loss is deterministic, and confirmed in experiments over this data (not shown for the sake of conciseness).

\begin{figure}
  \centering
  \subfloat[LML Loss.]{
      \includegraphics[width=0.475\textwidth]{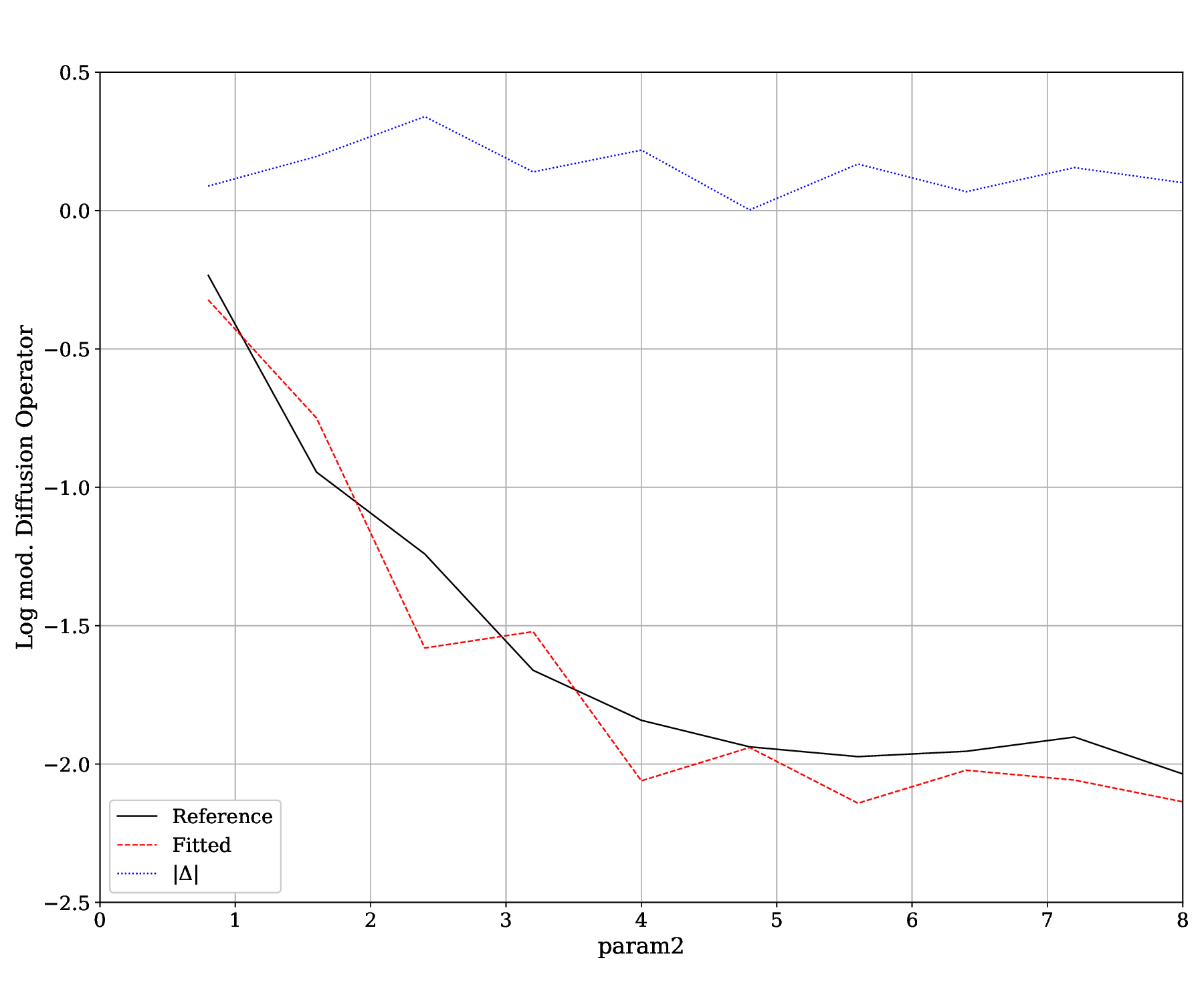}}
  \hfill
  \subfloat[Diffusion Loss.]{
      \includegraphics[width=0.475\textwidth]{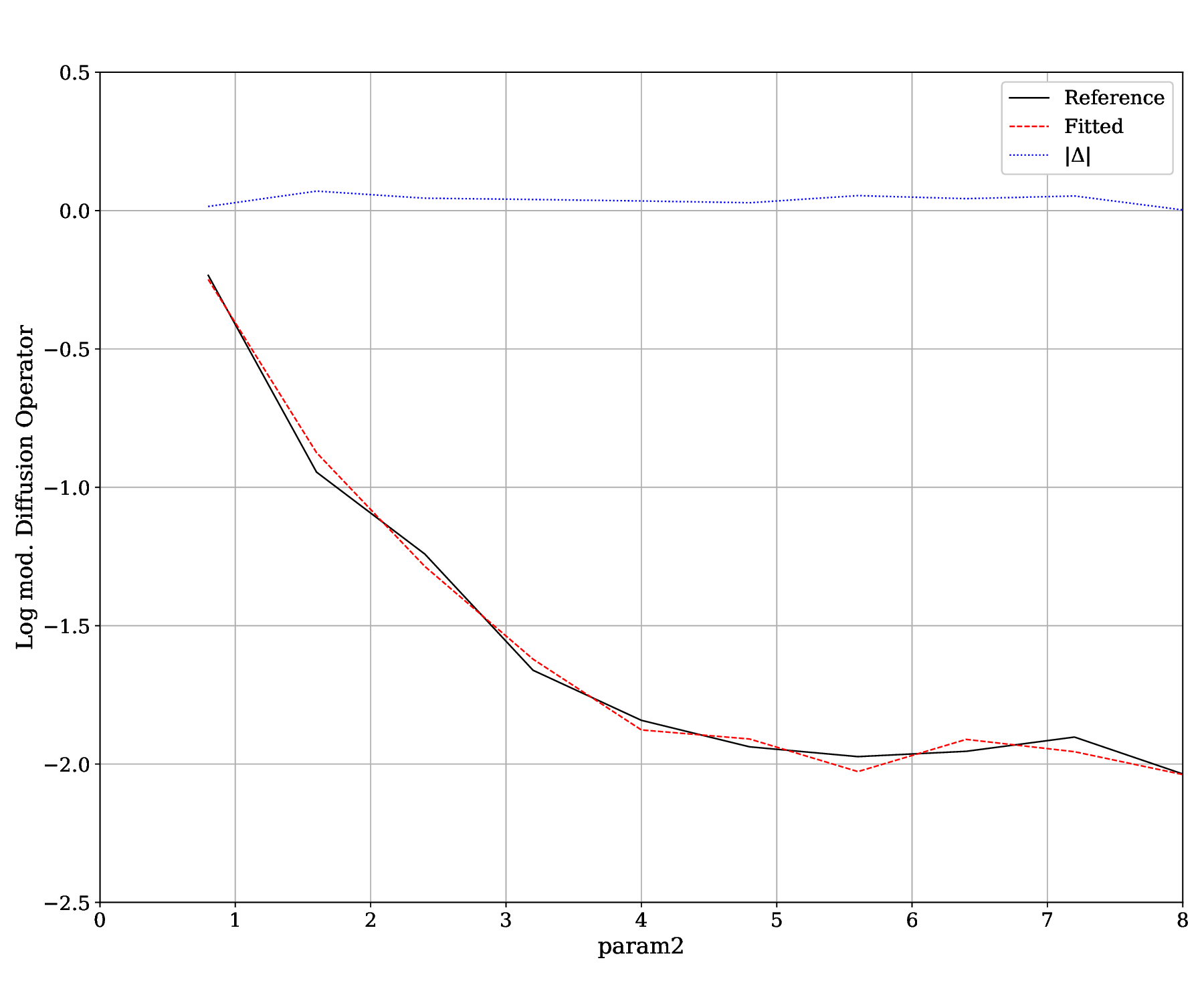}}
  \caption{Diffusion operators sliced at $(param1, param3) = (14, 0.7)$, showing the reference training values and the trained data results, both obtained with the SE kernel.}
  \label{slices_lapl}
\end{figure}

A comparison of the final diffusion operator at the same slice of Fig.\ \ref{result_lapl_loss} is shown in Fig.\ \ref{slices_lapl}.
As expected, the diffusion-loss method successfully approximates the entropy of the trained model towards the reference training distribution, decreasing the overfitting tendency as discussed.
The largest differences are observed at the highest curvature of the training data field, as also intuitively expected.

\section{Concluding Remarks}
A method to decrease the entropy contained in overfitting is proposed for data regression.
The method is based on the minimisation of the loss of a modified Laplace operator and tests for oscillation in the interior of the training mesh cells.
A true label is provided by the modified Laplace operator applied to the original training nodes.
This label acts as a measure of the ``entropy-in-features'' contained in the training data.
A \textit{trained} model must bound the amount of extra features it determines based on the minimisation of the loss with respect to this ``entropy-in-features''.
The testing points must be created on a staggered fashion to be able to gauge any excess entropy in the interior of the mesh cells.

With the proposed method, the model can be trained on all available training points.
This approach does not need to split the data into a test set since testing is provided by the separate but hierarchically meaningful diffusion metric.
The proposed method was shown to successfully reduce the overfitting on a three-dimensional example with fast change of properties in the operating envelope of the modelled system.

The author expects the proposed method to be generalisable to other machine learning or dataset strategies other than the framework herein presented.
\begin{itemize}
\item The method is demonstrated on a Gaussian Process regression context, but it should be generally applicable to other machine learning techniques in which the training is based on a set of objective functions constructed upon losses of metrics.
\item The current dataset is provided in mesh format, which facilitates the computation of finite Laplace operators.
However, the method is extensible to any other dataset in which the diffusion operator can be computed, for instance, unstructured meshes and graph/spectral Laplacians -- these structures would likely benefit from the edge-centred staggering as in a \oned case.
\item For datasets with noise, a true label is undefined in the sense presented in this work. However, a testing diffusion computed on staggered points can still be considered as an objective function to be independently minimised, enforcing a global reduction of ``entropy-in-features'' without a true label.
\end{itemize}

Some open questions that merit further research involve the reduction of the number of centre-crossing diagonals for large-dimensional problems.
The proposed method also involves Laplace operators only computed at the interior nodes of the mesh.
Biased finite differences at boundaries are definable and could be an extension of the current method for problems in which careful treatment of boundaries are in order.

Yet one other interesting possibility with the method is that, instead of using the Laplace operators as a loss metric, they could be used to directly reconstruct properties at the cell centroids.
These reconstructed values at the cell centres can be considered as the testing space and the loss of the properties at these points used for model training.
This approach may be considered too forceful by providing an artificial solution to train the model upon, but might have its merit on specific cases as it does not require the computation of the diffusion operators at every optimisation step.

In conclusion, the Laplace operator with its measure of the diffusion of a system provides a powerful tool with different facets to many different disciplines.
One additional role as a true label and loss for prevention of overfitting in data regression is here proposed. 
The method seems to show potential for a broader range of machine learning techniques.
This method does not replace manual work nor human interpretation to find ranges of good settings but lends a surer hand in the process by bringing more robustness to the process of, and better aim at, finding reasonable hyperparameters for model training.

\acks{The author thanks his ex-colleagues Yannick Bunk and Lorenzo Petrozzino for opening up to him the path of GPR for aeronautical engineering applications. The author has received no funding and has no conflict of interest to declare.}

\appendix
\section{} \label{appA}
This appendix provides a brief overview of the implementation in Python of the modified Laplace operator in \threed.
The implementation is straightforward with the use of the slicing feature for dealing with arrays in Python.
The training and the staggered meshes must be available for that, so first we define them.
\begin{lstlisting}[language=Python, basicstyle=\tiny]
	# Training mesh 'meshpts' is stacked
	#_ XX, YY, and ZZ are the unique values in each direction
	XX = np.unique( meshpts[0] )
	YY = np.unique( meshpts[1] )
	ZZ = np.unique( meshpts[2] )

	#_ Put into a meshgrid format so we can create the staggered mesh
	II, JJ, KK = np.meshgrid(XX, YY, ZZ, indexing='ij')
	
	# Staggered mesh coordinates, a simple average in each direction
	staggeredmesh_I = ( II[:-1, :-1, :-1] + II[1:, :-1, :-1]
	                  + II[:-1,1:  , :-1] + II[1:,1:,   :-1]
	                  + II[:-1, :-1,1:  ] + II[1:, :-1,1:  ]
	                  + II[:-1,1:,  1:  ] + II[1:,1:  ,1:  ]) / 8
	
	staggeredmesh_J = ( JJ[:-1, :-1, :-1] + JJ[1:, :-1, :-1]
	                  + JJ[:-1,1:  , :-1] + JJ[1:,1:  , :-1]
	                  + JJ[:-1, :-1,1:  ] + JJ[1:, :-1,1:  ]
	                  + JJ[:-1,1:  ,1:  ] + JJ[1:,1:  ,1:  ]) / 8
	
	staggeredmesh_K = ( KK[:-1, :-1, :-1] + KK[1:, :-1, :-1]
	                  + KK[:-1,1:  , :-1] + KK[1:,1:  , :-1]
	                  + KK[:-1, :-1,1:  ] + KK[1:, :-1,1:  ]
	                  + KK[:-1,1:  ,1:  ] + KK[1:,1:  ,1:  ]) / 8
	
	# Stack the mesh columnwise
	staggeredpts = np.c_[staggeredmesh_I.ravel(), staggeredmesh_J.ravel(),
	                     staggeredmesh_K.ravel()]
\end{lstlisting}

Given the coordinates of the two meshes, we use a model to predict the values at these points, termed \textit{f\_orig} for the training mesh nodes, and \textit{f\_stag} for the staggered nodes.
Now the computation of the Laplace operators can be executed as follows.

For the training mesh, only the training mesh nodes are used:
\begin{lstlisting}[language=Python, basicstyle=\tiny]
	dsf_dD1s = np.abs(f_orig[2:  ,2:  ,2:  ] + f_orig[ :-2, :-2, :-2]
	                 - 2.*f_orig[1:-1,1:-1,1:-1]) \
	               / (f_orig[2:  ,2:  ,2:  ] + f_orig[ :-2, :-2, :-2]
	                 + 2.*f_orig[1:-1,1:-1,1:-1])
	dsf_dD2s = np.abs(f_orig[ :-2,2:  ,2:  ] + f_orig[2:  , :-2, :-2]
	                 - 2.*f_orig[1:-1,1:-1,1:-1]) \
	               / (f_orig[ :-2,2:  ,2:  ] + f_orig[2:  , :-2, :-2]
	                 + 2.*f_orig[1:-1,1:-1,1:-1])
	dsf_dD3s = np.abs(f_orig[2:  , :-2,2:  ] + f_orig[ :-2,2:  , :-2]
	                 - 2.*f_orig[1:-1,1:-1,1:-1]) \
	               / (f_orig[2:  , :-2,2:  ] + f_orig[ :-2,2:  , :-2]
	                 + 2.*f_orig[1:-1,1:-1,1:-1])
	dsf_dD4s = np.abs(f_orig[2:  ,2:  , :-2] + f_orig[ :-2, :-2,2:  ]
	                 - 2.*f_orig[1:-1,1:-1,1:-1]) \
	               / (f_orig[2:  ,2:  , :-2] + f_orig[ :-2, :-2,2:  ]
	                 + 2.*f_orig[1:-1,1:-1,1:-1])
	            
	return (dsf_dD1s + dsf_dD2s + dsf_dD3s + dsf_dD4s) / delta
\end{lstlisting}
whereas for the staggered mesh, both training and staggered nodes must be used:
\begin{lstlisting}[language=Python, basicstyle=\tiny]
	dsf_dD1s = np.abs(f_orig[2:  ,2:  ,2:  ] + f_orig[ :-2, :-2, :-2]
	                - f_stag[1:  ,1:  ,1:  ] - f_stag[ :-1, :-1, :-1]) \
	               / (f_orig[2:  ,2:  ,2:  ] + f_orig[ :-2, :-2, :-2]
	                + f_stag[1:  ,1:  ,1:  ] + f_stag[ :-1, :-1, :-1])
	dsf_dD2s = np.abs(f_orig[ :-2,2:  ,2:  ] + f_orig[2:  , :-2, :-2]
	                - f_stag[ :-1,1:  ,1:  ] - f_stag[1:  , :-1, :-1]) \
	               / (f_orig[ :-2,2:  ,2:  ] + f_orig[2:  , :-2, :-2]
	                + f_stag[ :-1,1:  ,1:  ] + f_stag[1:  , :-1, :-1])
	dsf_dD3s = np.abs(f_orig[2:  , :-2,2:  ] + f_orig[ :-2,2:  , :-2]
	                - f_stag[1:  , :-1,1:  ] - f_stag[ :-1,1:  , :-1]) \
	               / (f_orig[2:  , :-2,2:  ] + f_orig[ :-2,2:  , :-2]
	                + f_stag[1:  , :-1,1:  ] + f_stag[ :-1,1:  , :-1])
	dsf_dD4s = np.abs(f_orig[2:  ,2:  , :-2] + f_orig[ :-2, :-2,2:  ]
	                - f_stag[1:  ,1:  , :-1] - f_stag[ :-1, :-1,1:  ]) \
	               / (f_orig[2:  ,2:  , :-2] + f_orig[ :-2, :-2,2:  ]
	                + f_stag[1:  ,1:  , :-1] + f_stag[ :-1, :-1,1:  ])
	
	return (dsf_dD1s + dsf_dD2s + dsf_dD3s + dsf_dD4s) / (3.*delta)
\end{lstlisting}
The referenced mesh spacing \textit{delta} is implemented as such
\begin{lstlisting}[language=Python, basicstyle=\tiny]
    if f_orig is not f_stag:
        # We have a staggered mesh with the original mesh at the centre/corners
        #_ We compute the Laplacian with a 5-point stencil
        grid_spacing = 0.5
    else:
        # This is the original training mesh processing, fstag = f_orig
        #_ We apply a 3-point stencil to compute the Laplacian
        grid_spacing = 1.

    delta = 3. * grid_spacing**2.
\end{lstlisting}

The complete code and datasets are available in \url{https://github.com/ebiga/gpr_aniso_trials}.

\vskip 0.2in
\bibliography{biblio}

\end{document}